\documentclass[pdflatex,default]{sn-jnl}

\usepackage{tabularx,algpseudocode,placeins,colortbl}
\algrenewcommand{\algorithmicrequire}{\textbf{Input:}}
\algrenewcommand{\algorithmicensure}{\textbf{Output:}}
\algrenewcommand{\algorithmicindent}{1em}

\jyear{2026}%

\theoremstyle{thmstyleone}%
\theoremstyle{thmstyletwo}%

\theoremstyle{thmstylethree}%

\makeatletter
\let\muvirl@headings\ps@headings
\def\ps@headings{%
  \muvirl@headings
  \def\@evenhead{\hbox to \hsize{\headerfont\thepage\qquad\rightmark\hfill}}%
  \def\@oddhead{\hbox to \hsize{\headerfont\hfill\leftmark\qquad\thepage}}%
}
\let\muvirl@titlepage\ps@titlepage
\def\ps@titlepage{%
  \muvirl@titlepage
  \let\@oddhead\@empty\let\@evenhead\@empty
}
\makeatother

\begin{document}

\title[MuViRL: Multimodal View-Invariant Representation Learning]{MuViRL: Multimodal View-Invariant Visual Speech Representation Learning via Calibrated View Aggregation}


\author[1,2]{\fnm{Fei} \sur{Su}}
\author[3]{\fnm{Kai} \sur{Qiu}}
\author*[2,1]{\fnm{Juan} \sur{Liu}}



\affil[1]{\orgname{School of Computer Science, Wuhan University, Wuhan, China}}

\affil[2]{\orgname{School of Artificial Intelligence, Wuhan University, Wuhan, China}}

\affil[3]{\orgname{Xiaomi EV Technology Co., Ltd., Beijing, China}}


\abstract{Multimodal representation learning combines complementary signals to capture shared semantic information. For dynamic scenes, these representations need to remain consistent across views while preserving distinctions between temporal states. In visual speech representation learning, view variation changes the visibility and reliability of articulation cues, making it difficult to learn such representations from synchronized audio and video. To address this challenge, we propose MuViRL, a self-supervised framework for multimodal view-invariant visual speech representation learning via calibrated view aggregation. The framework combines synchronized 3D Gaussian view synthesis with multimodal self-supervision to learn consistent representations from real and synthesized data. A multimodal decision model learns relative view utility from teacher prediction losses, and its calibrated scores guide the aggregation of complementary information. Aggregation-to-Single Distillation transfers the aggregated predictions to a single branch. Real-synthetic alignment matches synchronized windows across views while preserving their temporal relations. Experiments on real and synthesized multiview data demonstrate that MuViRL improves multimodal representation consistency and visual speech recognition performance across views.}




\maketitle

\section{Introduction}\label{sec:introduction}
Multimodal representation learning combines complementary observations to capture semantic information shared across modalities. Self-supervised MultiModal Versatile networks~\cite{alayrac2020mmv} and ImageBind~\cite{girdhar2023imagebind} use naturally paired signals to learn transferable representations. In dynamic scenes, these representations must also preserve changes over time. Visual speech representation learning draws on synchronized sound and lip motion to capture evolving articulation. The same articulation can appear different across views, while distinct states may differ only in subtle lip movements. Learning useful representations requires separating variation caused by viewpoint from changes that distinguish articulation states.

Multiview learning provides mechanisms for both view invariance and the fusion of complementary information. MVContrast~\cite{wang2023mvcontrast} and ViewCLR~\cite{das2023viewclr} learn consistency across object views and generated latent video views, respectively. For heterogeneous inputs, robust multiview learning~\cite{xu2025rml} combines sample-level attention with alignment under simulated perturbations. Decomposition and composition~\cite{wang2026skeleton} jointly train fused and individual modality features for skeleton-based action learning. Applying these ideas to visual speech requires correspondence within an utterance. Two segments can share the same utterance identity yet depict different articulation states. Agreement between whole-sequence representations does not establish which segments correspond across views or explicitly constrain their temporal relations.

Audiovisual pretraining supplies temporal supervision within a recording. AV-HuBERT~\cite{shi2022avhubert} predicts masked multimodal units, while RAVen~\cite{haliassos2023raven} and BRAVEn~\cite{haliassos2024braven} learn through prediction across modalities. CAV-MAE Sync~\cite{araujo2025sync} models fine-grained correspondence between sound and video. These approaches associate audio with visible events along an observed sequence. However, monocular datasets such as LRS3~\cite{afouras2018lrs3} and VoxCeleb2~\cite{chung2018vox} lack synchronized visual observations of the same articulation from different viewpoints. Synchronization between audio and video does not directly supervise how that articulation should be represented from another view.

Synchronized view synthesis can supply these observations, but matching timestamps does not ensure that each view retains useful articulation cues. Occlusion can hide mouth structures, and reconstruction errors can distort their shape or motion. Giving such views equal influence can propagate these errors through aggregation and alignment. Research on prediction quality~\cite{zhang2023qmf} and uncertainty~\cite{gao2024eau} addresses unreliable inputs in multimodal fusion. For synthesized visual speech, image sharpness alone cannot determine whether a view preserves the articulation needed for prediction. A sharp rendering may still distort a lip contour, while a valid perspective change may lower pixel similarity. The challenge is to assess predictive usefulness at corresponding times and use it to guide learning across views while preserving temporal distinctions.

To address these challenges, we propose MuViRL, a self-supervised framework for multimodal view-invariant visual speech representation learning via Calibrated View Aggregation. Region-Focused 3D Gaussian Splatting (3DGS) View Synthesis uses dynamic head reconstruction~\cite{qian2024avatars,stavatar2026} to generate synchronized observations, with fitting focused on articulatory structures. A frozen teacher derives unit targets from the original audio and video. We use these targets to supervise corresponding sequences and measure each view's predictive usefulness.

Calibrated View Aggregation uses a multimodal decision model to assess synchronized lip grids under a language rubric for articulation. Teacher prediction losses provide grade and ranking supervision, and temperature calibration adjusts the resulting grade probabilities. Calibrated scores and geometric correspondence guide Adaptive Weighting, linking each view's contribution to its usefulness for multimodal prediction.

Aggregation-to-Single Distillation (Agg2S) transfers aggregated predictions to a single branch that shares the encoder and unit targets. Temporal Real--Synthetic Representation Alignment (RSA) matches synchronized windows and their temporal relations. Reliability weighting reduces the influence of unreliable synthesized features during alignment. The encoder processes an individual video at inference without view aggregation or decision scoring.

The main contributions are as follows:
\begin{enumerate}
\item[1)] We propose MuViRL for multimodal view-invariant visual speech representation learning. Synchronized 3DGS views and shared audiovisual targets extend self-supervision across viewpoints while retaining temporal correspondence.
\item[2)] We introduce Calibrated View Aggregation to learn view utility from a frozen teacher's prediction losses. Calibrated decision scores, geometric correspondence, and Adaptive Weighting jointly determine each view's contribution.
\item[3)] We combine Agg2S with Temporal RSA to transfer aggregated predictions to individual views and align synchronized representations and their temporal relations. The learned encoder operates on a single video at inference.
\item[4)] We evaluate MuViRL on real and synthesized multiview data through representation analyses and recognition experiments. The results demonstrate improved correspondence across viewpoints and modalities while preserving discriminative information.
\end{enumerate}

\section{Related Work}\label{sec:related}
\subsection{Multimodal Representation Learning}
Multimodal representation learning uses correspondence between inputs to learn shared content and retain complementary information. Shared embeddings in MultiModal Versatile networks~\cite{alayrac2020mmv} and ImageBind~\cite{girdhar2023imagebind} connect modalities through paired observations. This principle supports multilingual pretraining~\cite{zhang2023multilingual} and image--text retrieval~\cite{lu2023retrieval}, and supplies semantic correspondences for object navigation~\cite{yuan2025navigation}. Reconstruction and fusion complement semantic matching by retaining cues that can help recover or interpret another modality. MultiMAE~\cite{bachmann2022multimae} learns to reconstruct masked inputs across modalities, whereas attention bottlenecks~\cite{nagrani2021bottlenecks} restrict feature exchange during fusion. Decomposition and composition~\cite{wang2026skeleton} supervise both fused and individual representations, linking the information retained by each for action learning.

For audiovisual sequences, synchronization provides correspondence at the level of temporal states. AV-HuBERT~\cite{shi2022avhubert} uses shared unit targets, while RAVen~\cite{haliassos2023raven} and BRAVEn~\cite{haliassos2024braven} learn through feature prediction. CAV-MAE Sync~\cite{araujo2025sync} and AlignVSR~\cite{liu2025alignvsr} explicitly align local audio and visual information. Temporal attention~\cite{kim2024temporal} further models their dependencies. Synthetic supervision in SynthVSR~\cite{liu2023synthvsr} expands recognition training data through speech-driven lip animation. Although these methods exploit the relationship between sound and visible articulation, pairing modalities within a recording does not supply observations of the same articulation from different viewpoints. MuViRL extends this supervision to synchronized original and synthesized views, using shared audiovisual targets to preserve articulation distinctions across views.

\subsection{Self-Supervised Multiview Learning}
The choice of paired views determines which variations a self-supervised objective suppresses. Contrastive Multiview Coding~\cite{tian2020cmc} extracts shared information from sensory channels. For physical viewpoint variation, MVContrast~\cite{wang2023mvcontrast} learns associations within views, across views, and between aggregated shape representations. ViewCLR~\cite{das2023viewclr} generates latent viewpoints for video learning. Robust multiview learning~\cite{xu2025rml} targets imperfect heterogeneous inputs, combining sample attention with alignment under simulated noise and missing information. Beyond matching paired views, aggregation determines how their complementary information is retained. Deep Sets~\cite{zaheer2017deepsets} formalizes invariance to input order, while Set Transformer~\cite{lee2019settransformer} models interactions among unordered inputs.

For dynamic sequences, aggregation must also preserve the information needed to distinguish temporal states. Supervised multiview lipreading combines synchronized streams~\cite{petridis2017multiview} or shares information through multitask learning~\cite{han2017multitask}. Temporal prediction captures dependencies within a sequence. Model compression~\cite{bucilua2006compression} transfers ensemble predictions to a single model, while similarity-preserving distillation~\cite{tung2019similarity} transfers relations among representations. However, agreement between whole sequences does not specify which articulation states should match across views. MuViRL combines shared target prediction and Agg2S with Temporal RSA to align synchronized windows and their relations. This preserves temporal correspondence during multiview learning while supporting inference from a single video.

\subsection{Calibrated Multimodal Decisions}
Reliability estimation guides fusion when inputs differ in quality. QMF~\cite{zhang2023qmf} relates dynamic fusion weights to prediction losses, while EAU~\cite{gao2024eau} uses aleatoric uncertainty to stabilize unimodal features before fusion. For ocular recognition, image quality estimation~\cite{yan2024ocular} weights fusion using embedding robustness. Geometric alignment also matters for fusion: cooperative segmentation of LiDAR and image inputs~\cite{guan2025lidar} preserves semantic and geometric correspondence during feature interaction. For synthesized speech views, reliability assessment must distinguish reconstruction defects from valid appearance changes across viewpoints.

Language-guided assessment specifies which visual structures should be preserved. MLLM-as-a-Judge~\cite{chen2024mllmjudge} examines scoring and ranking, while Q-Align~\cite{wu2024qalign} and DeQA-Score~\cite{you2025deqa} express image quality through ordered levels and score distributions. SmolVLM~\cite{marafioti2025smolvlm} provides compact image--language encoding. Pairwise logistic ranking~\cite{burges2005ranknet} learns relative order, whereas temperature scaling~\cite{guo2017calibration} and Dirichlet calibration~\cite{kull2019dirichlet} adjust probabilities. For synthesized speech views, general quality scores may miss articulation cues needed for multimodal prediction. We therefore use a frozen teacher's prediction losses to supervise utility grades and rankings, then calibrate the grade probabilities. Calibrated View Aggregation combines these scores with geometric correspondence and learned feature weights.

\subsection{Dynamic 3D Gaussian View Synthesis}
View synthesis provides additional observations when recordings cover few head orientations. Pose augmentation for lipreading~\cite{cheng2020pose} uses a fitted 3D morphable model to generate training views for supervised recognition. Dynamic avatars model facial motion and appearance: FLAME~\cite{li2017flame} parameterizes facial shape and expression, and Neural Head Avatars~\cite{grassal2022nha} combines deformable geometry with learned appearance. With efficient rendering provided by 3D Gaussian Splatting~\cite{kerbl2023gs}, mesh binding in GaussianAvatars~\cite{qian2024avatars} and separate face and mouth deformations in TalkingGaussian~\cite{li2024talkinggaussian} preserve facial structure during animation. STAvatar~\cite{stavatar2026} further improves motion and occluded detail through soft binding and temporal density control.

Geometric consistency also determines whether synthesized observations can be aligned across views. Two-dimensional Gaussian splatting~\cite{huang2024gs} emphasizes surface accuracy, and VCR-GauS~\cite{chen2024vcr} couples depth and normal regularization. DepthSplat~\cite{xu2025depthsplat} connects depth estimation with Gaussian rendering. Robust cross-view consistency~\cite{zhao2024crossview} and video depth adaptation~\cite{xia2026depth} further constrain geometry across observations. Reconstruction fidelity, however, does not establish whether rendered articulation remains informative for audiovisual prediction. MuViRL uses region-focused fitting and rendering along the original timeline to preserve local articulation and audio correspondence. The synchronized views supervise multimodal representations, and their rendered geometry supplies reliability evidence for aggregation and alignment.

\section{Method}\label{sec:method}
We consider multimodal representation learning from audiovisual videos. Each video provides an original lip image sequence $X_0$ and a synchronized audio signal $X_a$. The objective is to learn representations that remain consistent across views while distinguishing different articulation states.

\begin{figure}[!t]
\centering\includegraphics[width=1\textwidth]{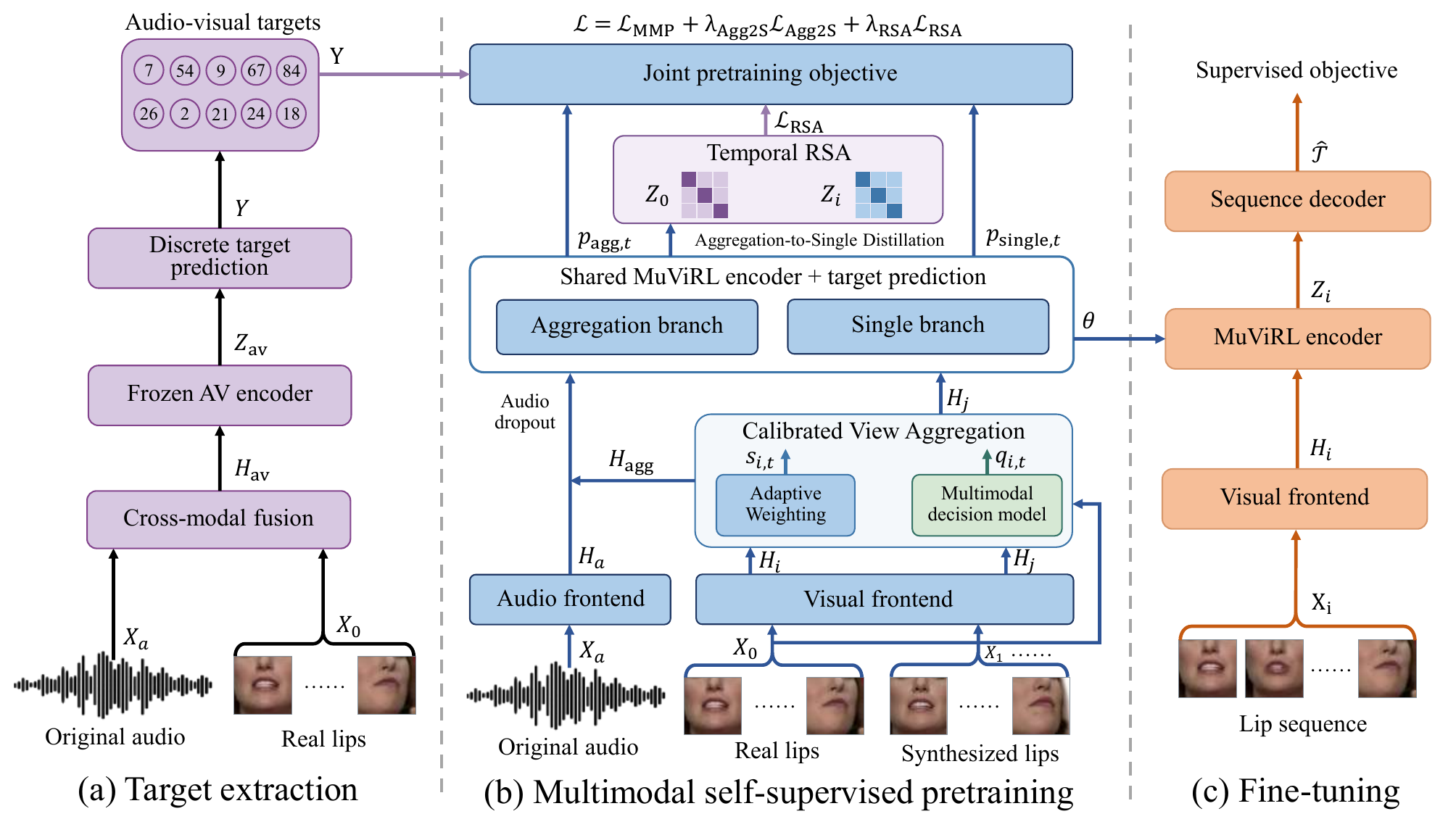}
\caption{Overview of the proposed framework. (a) Target extraction uses a frozen audiovisual teacher to generate shared unit targets $Y$. (b) Multimodal self-supervised pretraining uses Calibrated View Aggregation to combine visual features according to reliability $q_{i,t}$ and learned coefficients $s_{i,t}$. Aggregation-to-Single Distillation transfers aggregated predictions to the single branch, while MMP and Temporal RSA provide target supervision and temporal alignment. (c) Fine-tuning optimizes the pretrained encoder and a sequence decoder with transcript supervision.}\label{fig:fig_muvirl}
\end{figure}

An overview of our framework is shown in Fig.~\ref{fig:fig_muvirl}. Synchronized 3DGS synthesis provides corresponding views for multimodal masked prediction (MMP). Calibrated View Aggregation integrates features from these views. Aggregation-to-Single Distillation (Agg2S) transfers aggregated predictions to a single branch, while Temporal Real--Synthetic Representation Alignment (RSA) preserves synchronized states and their temporal relations.

\subsection{Region-Focused 3DGS View Synthesis}\label{sec:synthesis}
\begin{figure}[!t]
\centering\includegraphics[width=1\textwidth]{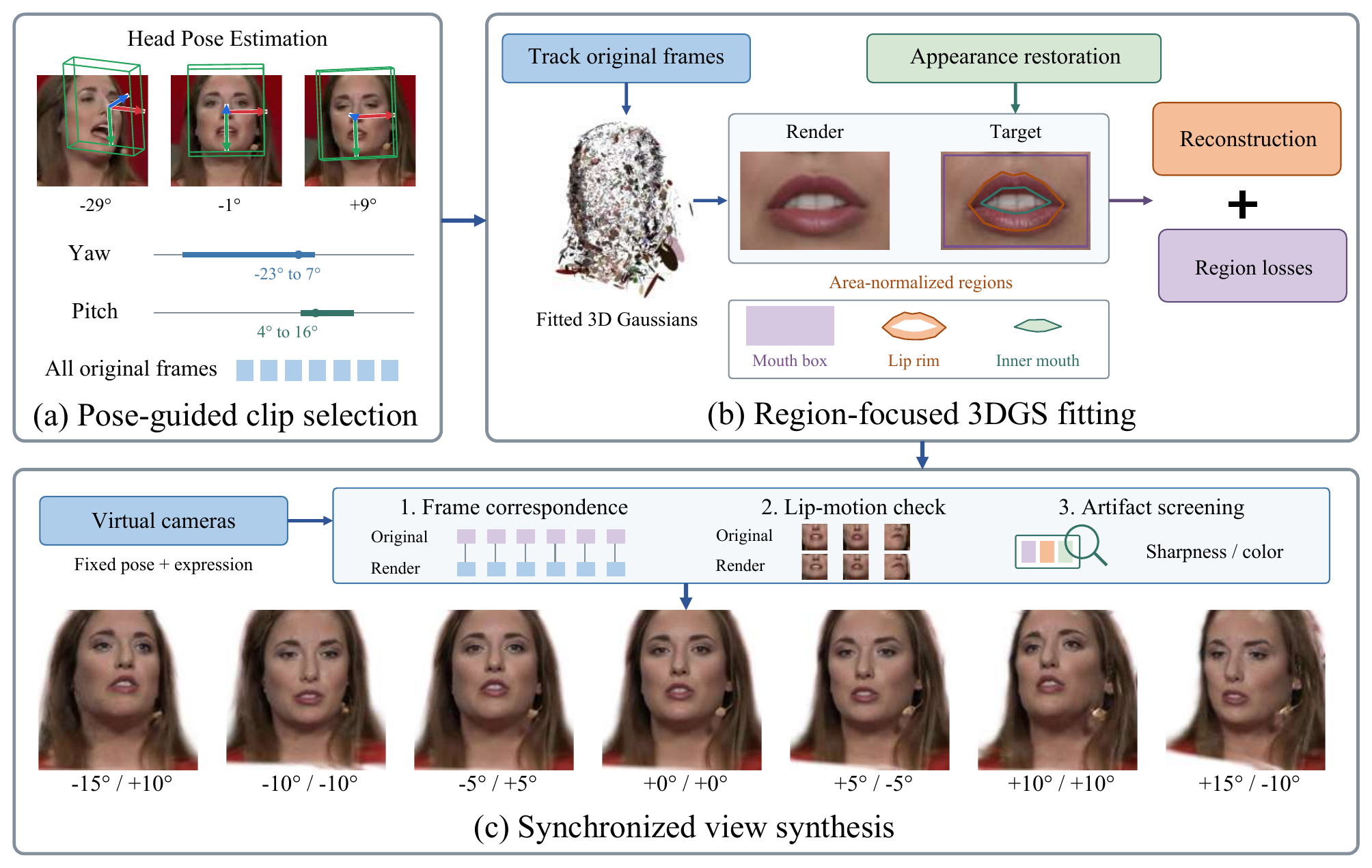}
\caption{Region-Focused 3DGS View Synthesis pipeline. (a) Pose-guided clip selection identifies clips with complementary yaw and pitch coverage from head pose trajectories. (b) Region-focused 3DGS fitting uses original motion and restored appearance for reconstruction, with errors averaged over the mouth box, lip rim, and inner mouth regions. (c) Synchronized view synthesis renders the same facial state through virtual cameras at each timestamp. Frame correspondence, lip motion, and rendering artifacts are then checked.}\label{fig:synthesis}
\end{figure}

We generate synchronized view sequences to expand viewpoint coverage while preserving lip structure and temporal correspondence. As illustrated in Fig.~\ref{fig:synthesis}, head rotations guide clip selection to expose complementary surfaces~\cite{grassal2022nha,hempel2022pose}. Foreground masks~\cite{lin2022rvm} and landmarks~\cite{zhou2023star} support tracking with VHAP\footnote{\url{https://github.com/ShenhanQian/VHAP}} and FLAME~\cite{li2017flame}. The recovered pose and expression remain fixed during appearance fitting.

After motion tracking, we restore the appearance targets using CodeFormer~\cite{zhou2022codeformer} while preserving the original frame indices. With the mesh binding strategy of STAvatar~\cite{stavatar2026}, a U-Net~\cite{ronneberger2015unet} predicts dynamic Gaussian attributes from appearance, UV positions, and frame displacements. Gaussian projection and compositing~\cite{kerbl2023gs} produce the renderings. The regional constraint $\mathcal L_{\mathrm{region}}=\sum_r\lambda_r\operatorname{MAE}(I_{\mathrm{render}}[\Omega_r],I_{\mathrm{target}}[\Omega_r])$ emphasizes the mouth box, lip rim, and inner mouth. Here, $\Omega_r$ selects region $r$, $\lambda_r$ weights its error, and MAE averages over its pixels and color channels. This constraint complements image reconstruction, SSIM~\cite{wang2004ssim}, LPIPS~\cite{zhang2018lpips}, and Gaussian regularization during avatar fitting.

Using the fitted avatar, we render each facial state from multiple virtual viewpoints at its original timestamp, retaining frame order and correspondence with the audio. Frame counts, lip motion, and rendering artifacts are checked before constructing training pairs. The resulting lip sequences are denoted by $X_i=\{X_{i,t}\}_{t=1}^{T}$, $i=1,\ldots,V$, where $V$ is the number of synthesized views and $t$ indexes the $T$ frames. View index zero denotes the original sequence. Rendered geometry supports reliability assessment. Section~S1.1 of the supplementary material (Online Resource~1) provides fitting and selection details, and Fig.~S1 illustrates the synthesized views and mouth reconstructions.

\begin{figure}[!t]
\centering\includegraphics[width=1\textwidth]{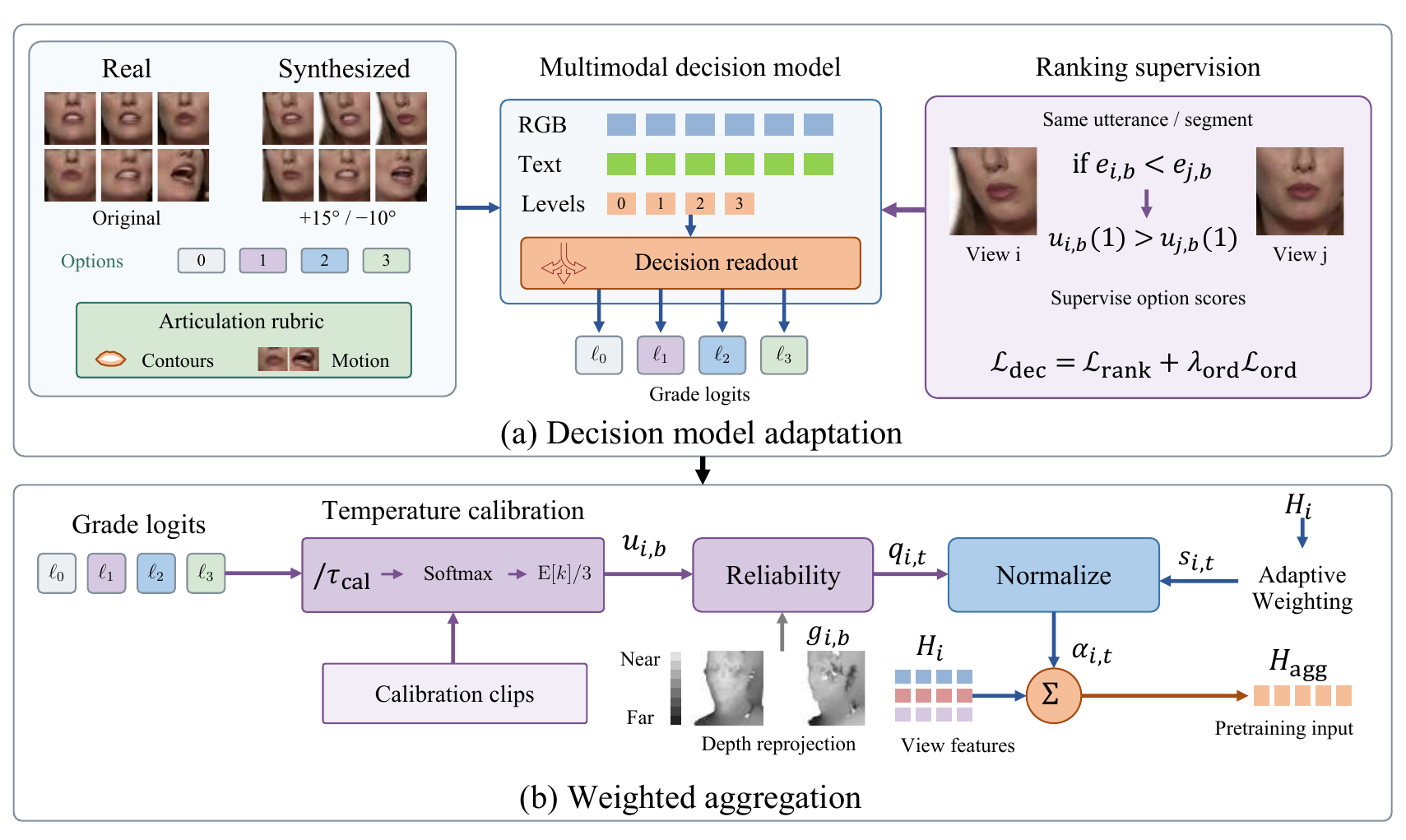}
\caption{Calibrated View Aggregation. (a) Decision model adaptation uses synchronized lip grids and a language rubric to predict articulation grades, supervised by teacher loss ordering and ordinal grades. (b) Weighted aggregation converts grade logits $\ell_k$ into calibrated utility $u_{i,b}$, then combines it with depth correspondence to obtain reliability $q_{i,t}$. Here, $u_{i,b}$ abbreviates $u_{i,b}(\tau_{\mathrm{cal}})$, and $\mathrm{E}[k]/3$ denotes the normalized expected grade. Reliability and learned coefficients $s_{i,t}$ determine weights $\alpha_{i,t}$ for aggregating view features into $H_{\mathrm{agg}}$.}\label{fig:aggregation}
\end{figure}

\subsection{Calibrated View Aggregation}\label{sec:aggregation}
To integrate complementary visual information for multimodal target prediction, we aggregate synchronized view features according to calibrated reliability. A multimodal decision model uses a language rubric to assess articulatory structures and motion in lip images affected by blur, occlusion, or rendering artifacts. Predictive utility supervises these assessments during adaptation. Temperature calibration adjusts the grade probabilities used to calculate reliability. As illustrated in Fig.~\ref{fig:aggregation}, calibrated reliability and Adaptive Weighting jointly determine each view's contribution. A shared visual frontend $F_v$ produces feature sequence $H_i$, with frame features $\mathbf h_{i,t}$, for aggregation.

\textbf{Multimodal decision model adaptation.} For each temporal segment $b$, paired original and candidate RGB grids contain frames at matching timestamps. The decision model $Q_\eta$ uses the Laya Vision\footnote{\url{https://github.com/r33drichards/laya-vision}} readout with SmolVLM~\cite{marafioti2025smolvlm} to jointly encode the grids, view angles, a language rubric, and four ordered options. The rubric specifies lip contours, opening changes, and visible teeth, allowing normal viewpoint variation and occlusion. Following Q-Align~\cite{wu2024qalign}, the four grades $k\in\{0,1,2,3\}$ range from insufficient to clearly preserved articulation. The model returns one logit $\ell_k$ per grade in a single forward pass. Applying softmax at temperature $\tau$ gives grade probabilities. Their expected grade, divided by three, defines the predictive utility score $u_{i,b}(\tau)\in[0,1]$~\cite{wu2024qalign}. A temperature of one gives the uncalibrated distribution, and $\tau_{\mathrm{cal}}$ gives the calibrated distribution.

Adaptation uses shared unit targets $Y=\{y_t\}_{t=1}^{T}$ generated by the frozen audiovisual teacher from the original pair $(X_0,X_a)$. Its visual prediction path evaluates each candidate at the same masked positions, producing segment loss $e_{i,b}$. Lower loss indicates greater predictive utility~\cite{zhang2023qmf}. For $e_{i,b}<e_{j,b}$, pairwise logistic ranking~\cite{burges2005ranknet} encourages $u_{i,b}(1)>u_{j,b}(1)$ by minimizing $\log(1+\exp((u_{j,b}(1)-u_{i,b}(1))/\tau_{\mathrm{rank}}))$, where $\tau_{\mathrm{rank}}>0$ controls score differences. Averaging the pairwise losses with weights derived from teacher loss differences gives $\mathcal L_{\mathrm{rank}}$. Candidate ranks define relative utility grades on the ordered readout scale within each segment, supplying soft targets for ordinal cross-entropy $\mathcal L_{\mathrm{ord}}$. After ordinal initialization, the readout and score head parameters $\eta$ are updated with $\mathcal L_{\mathrm{dec}}=\mathcal L_{\mathrm{rank}}+\lambda_{\mathrm{ord}}\mathcal L_{\mathrm{ord}}$. Temperature scaling~\cite{guo2017calibration} on independent calibration utterances determines $\tau_{\mathrm{cal}}$. We then freeze the decision model and cache its scores for student training. Supplementary Section~S2 details the prompts, grade targets, and calibration.

\textbf{Depth reprojection.} Rendered geometry supplies complementary evidence for correspondence~\cite{zhao2024crossview,chen2024vcr}. Mouth pixels are backprojected and transformed between views to compare depth and surface normals. Valid support, coverage, and residual errors determine a coefficient $g_{i,b}\in[0,1]$ using training statistics for each angle. The coefficient is one for geometrically consistent pairs and decreases as discrepancies increase. Views without rendered geometry also use one. Supplementary Section~S1.3 specifies the construction. Segment reliability is
\begin{equation}\label{eq:reliability}
q_{i,b}=q_{\min}+(1-q_{\min})u_{i,b}(\tau_{\mathrm{cal}})g_{i,b}.
\end{equation}
Here, $q_{\min}\in(0,1)$ sets a positive floor. Segment values map to frame weights $q_{i,t}$ by timestamp, with $q_{0,t}=1$ for the original view. Utility estimates the candidate's relative teacher grade. Geometry reduces the contribution of views with inconsistent correspondence.

\textbf{Weighted aggregation.} Adaptive Weighting learns each view's contribution from the selected group~\cite{wang2023mvcontrast,xu2025rml}. Let $\mathcal S$ denote the sampled aggregation subset and $\mathcal S_t$ its valid views at time $t$. A temporal convolution produces local descriptors. Each descriptor is concatenated with its difference from the group mean, then processed by a linear mapping, ReLU, and a two-layer MLP with a sigmoid output to obtain $s_{i,t}>0$. The final weights and aggregated features are
\begin{equation}\label{eq:aggregation}
\alpha_{i,t}=\frac{s_{i,t}\sqrt{q_{i,t}}}
{\sum_{j\in\mathcal S_t}s_{j,t}\sqrt{q_{j,t}}},
\qquad
\mathbf h_{\mathrm{agg},t}=\sum_{i\in\mathcal S_t}\alpha_{i,t}\mathbf h_{i,t}.
\end{equation}
The square root moderates reliability ratios. Temperature scaling changes the contrast among candidate scores and their weights relative to the original view when it is included. Cached $q_{i,t}$ remains fixed, whereas student gradients update Adaptive Weighting through $s_{i,t}$. Aggregation retains the timestamp and feature dimension, yielding sequence $H_{\mathrm{agg}}$. Jointly reordering views and their scores leaves the result unchanged~\cite{zaheer2017deepsets}.

\subsection{Aggregation-to-Single Distillation (Agg2S)}
\label{sec:distillation}
Agg2S transfers complementary information learned during multiview training to representations extracted from an individual video. The aggregation and single branches share the student encoder and predict the same unit targets. A separately sampled synthesized view $j$, which may also belong to the aggregation subset, supplies features $H_j$ to the single branch.

The audio frontend $F_a$ maps $X_a$ to synchronized features $H_a$. The aggregation branch combines $H_a$ and $H_{\mathrm{agg}}$ through concatenation, normalization, projection, masking, and temporal encoding. Audio dropout sets $H_a$ to zero on sampled updates. The single branch uses $H_j$ with audio fixed to zero. With student parameters $\theta$, the shared prediction path $F_\theta$ produces $p_{\mathrm{agg},t}(\tau)$ and $p_{\mathrm{single},t}(\tau)$ by applying softmax to logits divided by temperature $\tau$. Both branches predict targets at the same valid masked positions $\mathcal M$.

Agg2S matches their predictive distributions using Kullback--Leibler (KL) divergence~\cite{kim2021comparing}:
\begin{equation}\label{eq:agg2s}
\mathcal L_{\mathrm{Agg2S}}=\frac{\tau_{\mathrm d}^2}{N_{\mathrm m}}
\sum_{t\in\mathcal M}\operatorname{KL}\!\left(
p_{\mathrm{agg},t}(\tau_{\mathrm d})\,\middle\|\,
 p_{\mathrm{single},t}(\tau_{\mathrm d})\right).
\end{equation}
Here, $\tau_{\mathrm d}>0$ is the distillation temperature and $N_{\mathrm m}$ counts shared valid masked positions. The temperature squared compensates for gradient rescaling. The aggregation distribution is held fixed in this loss, so gradients update the single prediction path through the shared parameters. In the same iteration, the aggregation branch learns from target prediction, while the single branch learns from both unit targets and predictions based on complementary views. At inference, the shared encoder extracts representations from an individual video without view aggregation or decision scoring.

\subsection{Reliability-Aware Real--Synthetic Alignment}\label{sec:alignment}
Alignment encourages consistent representations of the same articulation across original and synthesized views while preserving distinctions between successive states. Temporal RSA aligns synchronized windows and their relations, combining correspondence across views~\cite{wang2023mvcontrast,xu2025rml} with similarity supervision~\cite{tung2019similarity}. Calibrated reliability weights both constraints to reduce the influence of unreliable synthesized frames.

\textbf{Synchronized window alignment.} To align articulation at corresponding times, we encode visual sequences without feature masking to obtain $Z_0$ and $Z_i$. Pooling jointly valid frames within window $w$ and applying L2 normalization gives original vector $\mathbf z_{\mathrm{ref},i,w}$ and candidate vector $\mathbf z_{i,w}$. The original vector is pooled over the same timestamps as candidate $i$ and detached along the alignment path. Weight $a_{i,w}$ averages cached $q_{i,t}$ over these frames. Invalid windows are excluded, and uniform alignment uses $a_{i,w}=1$. The weighted cosine distance between corresponding windows is
\begin{equation}\label{eq:sync}
\mathcal L_{\mathrm{sync}}=
\frac{\sum_{i,w}a_{i,w}\left[1-\mathbf z_{\mathrm{ref},i,w}^{\top}\mathbf z_{i,w}\right]}
{\sum_{i,w}a_{i,w}}.
\end{equation}
The inner product measures similarity between normalized vectors. Normalization by the total weight preserves the loss scale across valid windows. Gradients update the synthesized representation to align corresponding states.

\textbf{Temporal relation alignment.} To preserve distinctions between states, we also match the similarities between windows within each sequence:
\begin{equation}\label{eq:relation}
\mathcal L_{\mathrm{relation}}=
\frac{\displaystyle\sum_{i,w,w'}a_{i,w}a_{i,w'}
\left[\mathbf z_{\mathrm{ref},i,w}^{\top}\mathbf z_{\mathrm{ref},i,w'}
-\mathbf z_{i,w}^{\top}\mathbf z_{i,w'}\right]^2}
{\displaystyle\sum_{i,w,w'}a_{i,w}a_{i,w'}}.
\end{equation}
Here, $w,w'$ identify windows of the same utterance. The squared difference compares temporal relations in the original and synthesized sequences, while $a_{i,w}a_{i,w'}$ weights both endpoints. Sums include valid pairs across the minibatch. The combined objective is
\begin{equation}\label{eq:rsa}
\mathcal L_{\mathrm{RSA}}=\beta_{\mathrm{sync}}\mathcal L_{\mathrm{sync}}
+\beta_{\mathrm{rel}}\mathcal L_{\mathrm{relation}}.
\end{equation}
Nonnegative $\beta_{\mathrm{sync}}$ and $\beta_{\mathrm{rel}}$ balance the two reliability weighted constraints on synchronized windows and temporal relations.

\begin{algorithm}[!t]
\caption{Self-supervised optimization of MuViRL}\label{alg:muvirl}
\small
\setlength{\baselineskip}{10.75pt}
\begin{algorithmic}
\Statex \textbf{Decision model adaptation}
\Require $\mathcal D_{\mathrm{tr}},\mathcal D_{\mathrm{cal}}$ and frozen teacher.
\Ensure Adapted $Q_\eta$ and $\tau_{\mathrm{cal}}$.
\State \textbf{Initialize:} $\eta\leftarrow\eta_0$.
\State Extract $Y$ and candidate losses $\{e_{i,b}\}$.
\State Form pairs with $e_{i,b}<e_{j,b}$.
\State Construct ordinal targets.
\State Initialize $Q_\eta$ with $\mathcal L_{\mathrm{ord}}$.
\For{$n_{\mathrm{dec}}=1,\ldots,N_{\mathrm{dec}}$}
    \State Sample candidate groups from $\mathcal D_{\mathrm{tr}}$.
    \State $\mathcal L_{\mathrm{dec}}\leftarrow\mathcal L_{\mathrm{rank}}+\lambda_{\mathrm{ord}}\mathcal L_{\mathrm{ord}}$.
    \State Update $\eta$ using $\nabla_\eta\mathcal L_{\mathrm{dec}}$.
\EndFor
\State Fit $\tau_{\mathrm{cal}}$ on $\mathcal D_{\mathrm{cal}}$ and freeze $Q_\eta$.
\Statex \textbf{Multimodal self-supervised pretraining}
\Require $\{X_i\},X_a,Y$, geometry, and $Q_\eta$.
\Ensure Pretrained student parameters $\theta$.
\State \textbf{Initialize:} $\theta\leftarrow\theta_0$. Freeze $F_v,F_a$.
\State Cache $\{q_{i,t}\}$ using Eq.~\eqref{eq:reliability}.
\For{$n_{\mathrm{train}}=1,\ldots,N_{\mathrm{train}}$}
    \State Sample a minibatch and views $\mathcal S,j$.
    \State Sample shared mask $\mathcal M$.
    \For{$i\in\mathcal S\cup\{0,j\}$}
        \State $H_i\leftarrow F_v(X_i)$.
    \EndFor
    \State $H_a\leftarrow F_a(X_a)$. Apply audio dropout.
    \State $\{\alpha_{i,t}\},H_{\mathrm{agg}}\leftarrow$ Eq.~\eqref{eq:aggregation}.
    \State $p_{\mathrm{agg}}\leftarrow F_\theta(H_{\mathrm{agg}},H_a;\mathcal M)$.
    \State $p_{\mathrm{single}}\leftarrow F_\theta(H_j,0;\mathcal M)$.
    \State Encode $Z_0,\{Z_i\}$ without masking.
    \State Evaluate $\mathcal L_{\mathrm{MMP}},\mathcal L_{\mathrm{Agg2S}},\mathcal L_{\mathrm{RSA}}$.
    \State Update $\theta$ using $\nabla_\theta\mathcal L$ [Eq.~\eqref{eq:total}].
\EndFor
\end{algorithmic}
\end{algorithm}

\subsection{Real--Synthetic Self-Supervised Pretraining}\label{sec:pretraining}
\textbf{Target extraction.} The frozen audiovisual teacher fuses each original audio and lip sequence into $H_{\mathrm{av}}$, encodes it as $Z_{\mathrm{av}}$, and selects the most probable unit at each timestamp. The resulting targets $Y$ also supervise synthesized sequences through their shared frame indices.

\textbf{Multimodal masked prediction.} At shared positions $\mathcal M$, projected features are replaced by features from other valid timestamps. The aggregation branch uses audiovisual context with audio dropout, and the single branch uses visual context~\cite{shi2022avhubert}. Their target cross-entropies form
\begin{equation}\label{eq:mmp}
\mathcal L_{\mathrm{MMP}}=\frac{\mathcal L_{\mathrm{agg}}+
\lambda_{\mathrm{single}}\mathcal L_{\mathrm{single}}}{1+\lambda_{\mathrm{single}}}.
\end{equation}
Here, $\lambda_{\mathrm{single}}\geq0$ controls the single branch contribution. Normalization preserves the combined loss scale.

\textbf{Overall objective.} Prediction, distillation, and alignment jointly optimize the student:
\begin{equation}\label{eq:total}
\mathcal L=\mathcal L_{\mathrm{MMP}}+
\lambda_{\mathrm{Agg2S}}\mathcal L_{\mathrm{Agg2S}}+
\lambda_{\mathrm{RSA}}\mathcal L_{\mathrm{RSA}}.
\end{equation}
The coefficients $\lambda_{\mathrm{Agg2S}}$ and $\lambda_{\mathrm{RSA}}$ balance distillation and alignment.

For recognition, a sequence decoder maps $Z_i$ to the predicted transcript $\widehat{\mathcal T}$ and is fine-tuned with the encoder using transcript supervision.

\textbf{Optimization.} Algorithm~\ref{alg:muvirl} separates decision adaptation from student updates. Training and calibration utterances are denoted by $\mathcal D_{\mathrm{tr}}$ and $\mathcal D_{\mathrm{cal}}$. Student parameters $\theta$ comprise the updated encoder, projection, and aggregation parameters. The initial parameters are $\eta_0,\theta_0$, and $N_{\mathrm{dec}},N_{\mathrm{train}}$ specify the respective numbers of updates.

\FloatBarrier
\section{Experiments}\label{sec:experiments}
\subsection{Experimental Setup}
\textbf{Datasets.} We evaluate our proposed framework on LRS3~\cite{afouras2018lrs3}, VoxCeleb2~\cite{chung2018vox}, TCD-TIMIT~\cite{harte2015tcd}, and OuluVS2~\cite{anina2015oulu}. LRS3 provides 433\,h of transcribed English speech. Following the AV-HuBERT~\cite{shi2022avhubert} protocol, we combine pretrain and trainval for training, reserve approximately 1\,h from trainval for validation, and use the official test set. We assess label efficiency using the 30\,h trainval subset. LRS3 and 1,326\,h of English VoxCeleb2~\cite{chung2018vox} data form a 1,759\,h collection used for large-scale pretraining. TCD-TIMIT contains 6,913 sentences from 62 speakers, recorded from frontal and $30^\circ$ viewpoints. OuluVS2 provides synchronized video at five fixed viewpoints ($0^\circ$, $30^\circ$, $45^\circ$, $60^\circ$, $90^\circ$). Both datasets support evaluation of robustness to view variation on real multiview data. ORL~\cite{orl}, Yale~\cite{yale}, Notting Hill~\cite{cao2015clustering}, and MEAD~\cite{wang2020mead} provide quality grade supervision for decision model fine-tuning. The first three provide static face crops with variations in pose, expression, illumination, and occlusion, while MEAD provides synchronized multiview facial sequences.

\textbf{Multiview data synthesis.} We fit dynamic 3DGS avatars~\cite{stavatar2026} with 1.5 million Gaussians for 300 epochs, followed by 30 refinement epochs. Original mouth pixels are retained in the reconstruction targets. Each selected original clip yields twelve synchronized sequences, with horizontal offsets at $5^\circ$ intervals within $[-25^\circ,25^\circ]$ and pitch offsets within $[-10^\circ,15^\circ]$. View-Augmented (VA) training draws 60\% of base samples from original clips and 40\% from synthesized clips. VA-30h, VA-433h, and VA-1759h denote the respective original data collections. Original and synthesized views of an utterance remain in the same dataset partition. Supplementary Section~S1 provides synthesis and sampling details, and Fig.~S1 shows synthesized inputs.

\textbf{Metrics.} We use accuracy, recall, and cosine similarity to evaluate representation learning and aggregation. Visual speech recognition (VSR) uses word error rate (WER), $\mathrm{WER}=(S+D+I)/N_{\mathrm{ref}}\times100\%$, where $S$, $D$, and $I$ count substitutions, deletions, and insertions, and $N_{\mathrm{ref}}$ counts reference words.

\subsection{Implementation Details}
\textbf{Backbone and initialization.} 
We initialize from the 5th-iteration AV-HuBERT model~\cite{shi2022avhubert}, which provides cluster pseudo-labels for the 6th iteration. Base and Large use 12 layers of dimension 768 and 24 layers of dimension 1,024, respectively, with a ResNet-18~\cite{he2016resnet} visual frontend. Preprocessing converts $96\times96$ lip crops to grayscale $88\times88$ inputs. Audio preprocessing follows the AV-HuBERT configuration. Experiments are conducted on NVIDIA GeForce RTX 4090 GPUs.

\textbf{Decision inputs and prompts.} We use the Laya Vision 201M checkpoint. We divide each clip into three chronological segments and sample up to six evenly spaced frames per segment. Synchronized original and candidate frames are arranged in separate $2\times3$ RGB grids. The text specifies frame order, view angles, the assessment criterion, and four ordered grades. The predictive utility instruction includes ``Judge how useful the second view is for recovering the articulation shown in the reference: changes in lip contour, opening and visible teeth.'' The rubric accepts normal viewpoint variation and occlusion. The structural variant assesses preservation under blur, occlusion, and rendering defects. The model returns four logits for either prompt in one forward pass. Grades range from zero to three, and temperature scaling followed by the normalized expected grade produces the utility score. Supplementary Section~S2.1 provides the complete prompts.

\textbf{Decision adaptation and calibration.} Vision and language backbones remain frozen while approximately 8.31M readout and score head parameters are updated. Ordinal initialization uses 300 steps at $10^{-4}$ with batch size eight, followed by $N_{\mathrm{dec}}=300$ ranking steps at $3\times10^{-5}$ using twelve synchronized candidates. We set $\lambda_{\mathrm{ord}}=0.1$ and $\tau_{\mathrm{rank}}=0.25$, cap normalized pair weights at three, and apply label smoothing of 0.05. For the quality grade experiment, MEAD and the auxiliary face datasets supply original inputs and three levels of blur or occlusion. We fine-tune the readout for 1,500 updates at $3\times10^{-5}$ with four ordered examples per update. Calibration and evaluation use separate utterances or identities, and pixelation is reserved for evaluation. With the decision model frozen, we fit $\tau_{\mathrm{cal}}$ by minimizing grade negative log-likelihood and cache scores. Reliability uses $q_{\min}=0.25$ and geometric coefficient one when rendered geometry is unavailable. Supplementary Section~S2 gives label construction and partition details.

\textbf{Self-supervised pretraining and fine-tuning.} MuViRL uses the full objective in Eq.~\ref{eq:total}, with teacher grade and ranking supervision followed by temperature calibration. Reliability weights guide both view aggregation and Temporal RSA. The MuViRL configurations in Table~\ref{tab:overall} use Large encoders pretrained on VA-433h or VA-1759h and fine-tuned on 433\,h of labeled data. Optimization and recognition decoding follow AV-HuBERT~\cite{shi2022avhubert}. OuluVS2 fine-tuning follows the protocol of Han et al.~\cite{han2017multitask}. Representation, distillation, and ablation comparisons share twenty LRS3 training and eight validation utterances, with 600 pretraining updates and 150 recognition fine-tuning updates. AdamW~\cite{loshchilov2019adamw} updates the final two Base encoder layers, projection, normalization, and Adaptive Weighting while frontend features and unit prototypes remain frozen. We set $\lambda_{\mathrm{single}}=2$, $\lambda_{\mathrm{Agg2S}}=0.1$, $\lambda_{\mathrm{RSA}}=0.5$, $\tau_{\mathrm d}=2$, and $\beta_{\mathrm{sync}}=\beta_{\mathrm{rel}}=1$. Supplementary Section~S1.2 gives optimization budgets, sampling, and evaluation settings.

\subsection{Overall Performance Comparison}\label{sec:overall}
To evaluate MuViRL's view-invariant multimodal representations, we compare its performance with representative supervised and self-supervised methods on LRS3. Table~\ref{tab:overall} presents results for Base and Large encoders to assess the effect of model capacity. MuViRL and AV-HuBERT are compared using the same Large encoder and VA training data.

\begin{table}[!t]
\caption{Overall performance on LRS3. The $0^\circ$ column uses original test data. Other columns add synthesized views up to the indicated yaw magnitude in both directions, with zero pitch offset. VA denotes pretraining with synthesized multiview data.}\label{tab:overall}
\centering\fontsize{8}{10}\selectfont\setlength{\tabcolsep}{0.5pt}
\begin{tabular*}{\textwidth}{@{\extracolsep{\fill}}lcccccccc@{}}
\toprule
\multirow{2}{*}{\textbf{Method}} & \multirow{2}{*}{\textbf{Encoder}} & \multirow{2}{*}{\textbf{Criterion}} & \multicolumn{2}{c}{\textbf{Training data (h)}} & \multicolumn{4}{c}{\textbf{WER (\%)} $\downarrow$}\\
\cmidrule(lr){4-5}\cmidrule(lr){6-9}
&&&Labeled&Unlabeled&$\mathbf{0^\circ}$&$\mathbf{5^\circ}$&$\mathbf{10^\circ}$&$\mathbf{15^\circ}$\\
\midrule
\rowcolor[gray]{0.9}\multicolumn{9}{c}{\textit{Supervised}}\\
CM-seq2seq~\cite{ma2021conformer}&Conformer&CTC+S2S&433&--&43.3&64.6&70.8&83.9\\
XMD~\cite{afouras2020asr}&CNN&CTC&433&--&68.8&74.1&78.6&82.0\\
MSR~\cite{xu2020msr}&RNN&S2S&433&--&57.8&66.4&72.5&77.6\\
Auto-AVSR~\cite{ma2023auto}&Conformer&CTC+S2S&818&--&33.0&51.5&64.8&73.0\\
\midrule
\rowcolor[gray]{0.9}\multicolumn{9}{c}{\textit{Self-supervised}}\\
CMA~\cite{kim2024temporal}&Transformer&S2S&433&1,759&32.5&40.4&50.5&73.8\\
RAVen~\cite{haliassos2023raven}&Transformer&CTC+S2S&433&1,759&27.8&35.9&49.1&52.3\\
AV-HuBERT-B~\cite{shi2022avhubert}&Transformer&S2S&433&1,759&34.8&45.7&66.2&72.1\\
AV-HuBERT-L~\cite{shi2022avhubert}&Transformer&S2S&433&1,759&28.2&36.3&43.6&50.5\\
AV-HuBERT-L (VA-433h)&Transformer&S2S&433&433&44.6&63.7&71.5&82.4\\
AV-HuBERT-L (VA-1759h)&Transformer&S2S&433&1,759&28.6&40.2&46.6&52.7\\
\midrule
\rowcolor[gray]{0.9}\multicolumn{9}{c}{\textit{Our Method}}\\
MuViRL-L (VA-433h)&Transformer&S2S&433&433&42.2&61.4&68.1&71.4\\
MuViRL-L (VA-1759h)&Transformer&S2S&433&1,759&\textbf{27.0}&\textbf{34.3}&\textbf{40.8}&\textbf{46.9}\\
\bottomrule
\end{tabular*}
\end{table}

\begin{table}[t]
\caption{Multimodal representation evaluation. Here, MuViRL denotes the configuration trained without decision or depth weighting. Values are means $\pm$ standard deviations over three pretraining seeds.}\label{tab:representations}
\centering\fontsize{8}{10}\selectfont\setlength{\tabcolsep}{1.6pt}\medmuskip=0mu
\begin{tabularx}{\textwidth}{@{}Xcccccc@{}}
\toprule
\multirow{4}{*}[-0.8ex]{Configuration}&\multicolumn{3}{c}{LRS3}&\multicolumn{3}{c}{OuluVS2}\\
\cmidrule(lr){2-4}\cmidrule(lr){5-7}
&\shortstack{Prediction\\accuracy\\(\%) $\uparrow$}&\shortstack{Cross-view\\Recall\\(\%) $\uparrow$}&\shortstack{Cross-modal\\Recall\\(\%) $\uparrow$}&\shortstack{Prediction\\accuracy\\(\%) $\uparrow$}&\shortstack{Cross-view\\Recall\\(\%) $\uparrow$}&\shortstack{Cross-modal\\Recall\\(\%) $\uparrow$}\\
\midrule
Baseline&$3.85\pm0.03$&$64.59\pm0.27$&$53.56\pm0.23$&$29.17\pm0.41$&$83.40\pm0.35$&$64.02\pm0.60$\\
Random view&$\mathbf{3.93}\pm0.06$&$65.32\pm0.34$&$53.93\pm0.28$&$32.84\pm0.48$&$85.68\pm0.51$&$67.23\pm1.08$\\
MuViRL&$3.91\pm0.05$&$\mathbf{67.64}\pm0.06$&$\mathbf{55.32}\pm0.09$&$\mathbf{35.09}\pm0.33$&$\mathbf{88.66}\pm0.22$&$\mathbf{68.60}\pm0.67$\\
\bottomrule
\end{tabularx}
\end{table}

With synthesized yaw offsets up to $\pm15^\circ$, MuViRL-L pretrained on VA-1759h achieves 46.9\% WER, outperforming CMA (73.8\%) and RAVen (52.3\%). Using the same Large encoder and VA-1759h training data, it achieves 27.0\% WER on the original test set and 46.9\% with the $\pm15^\circ$ range, compared with 28.6\% and 52.7\% for AV-HuBERT-L. The relative improvement is greater with synthesized views than on the original test set, averaging 10.9\% across the four settings. Improvements are also obtained with VA-433h, where MuViRL-L achieves 71.4\% WER with the $\pm15^\circ$ range, compared with 82.4\% for AV-HuBERT-L. The improvements under both pretraining settings, with encoder capacity and training data matched, indicate robustness to viewpoint variation.

\subsection{Evaluation of Multimodal View-Invariant Representations}\label{sec:representations}
We use AV-HuBERT as the baseline to assess view invariance and generalization. The qualitative and quantitative comparisons cover recognition on real multiview data, spatial responses, feature distributions, and temporal correspondence.

\begin{figure}[!t]
\centering\includegraphics[width=1\textwidth]{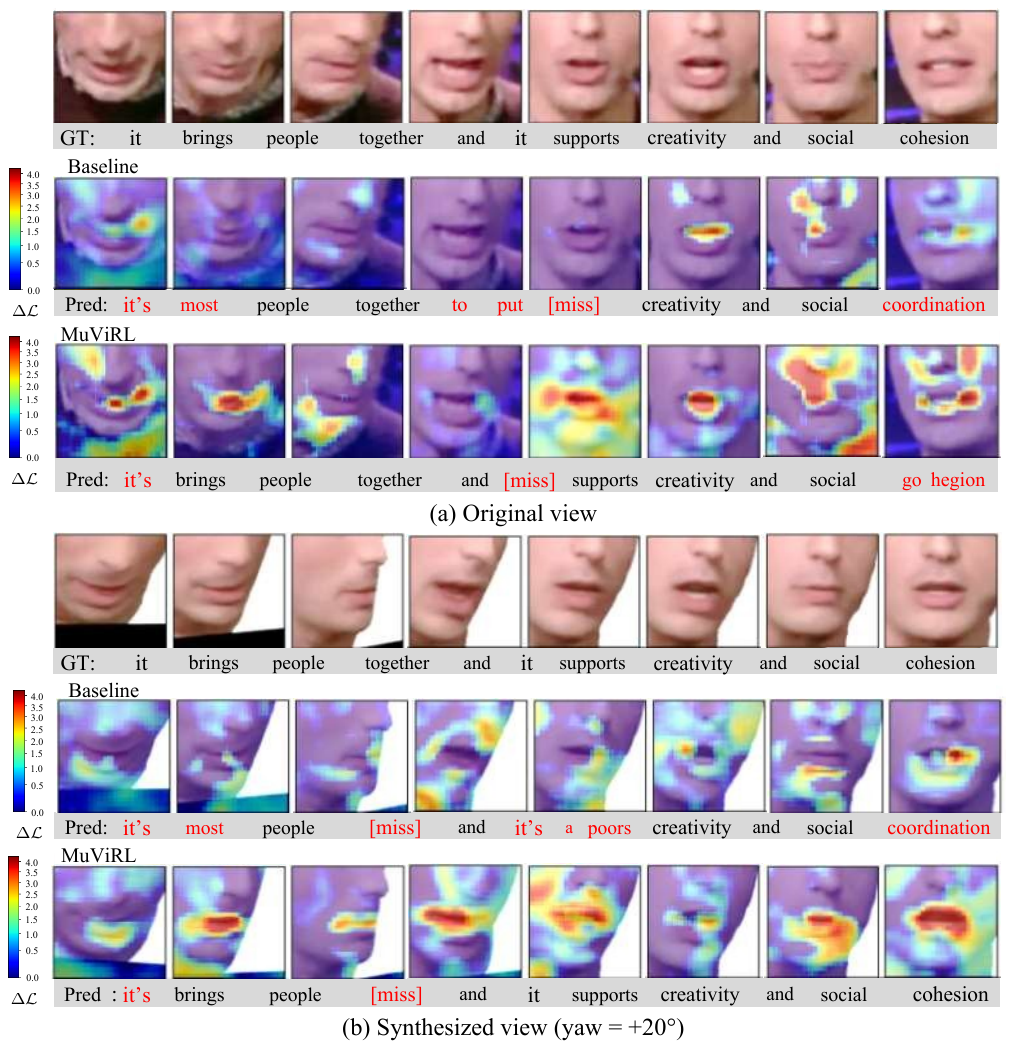}
\caption{Qualitative evaluation of view-invariant representations. We overlay sensitivity heatmaps on an LRS3 test sequence for (a) the original view and (b) a synthesized view with yaw $+20^\circ$, showing the contribution of different mouth regions to prediction. Colors range from blue (low sensitivity) to red (high sensitivity), according to the loss change $\Delta\mathcal L$ under local occlusion. Word predictions are aligned below the frames. Red words mark transcript mismatches, and \texttt{[miss]} denotes a deletion.}\label{fig:responses}
\end{figure}

\textbf{Qualitative evaluation.} We adopt a spatiotemporal occlusion sensitivity method similar to receptive-field visualization approaches~\cite{zeiler2014visualizing,zhou2015object} to identify regions contributing to prediction. As shown in Fig.~\ref{fig:responses}, the baseline's spatial responses vary across views, with additional prediction errors. In the illustrated sequence, MuViRL maintains sensitivity around the mouth opening and lip contours and makes fewer word prediction errors. These localized responses and predictions indicate preserved articulatory evidence across views.

\begin{figure}[!t]
\centering\includegraphics[width=1\textwidth]{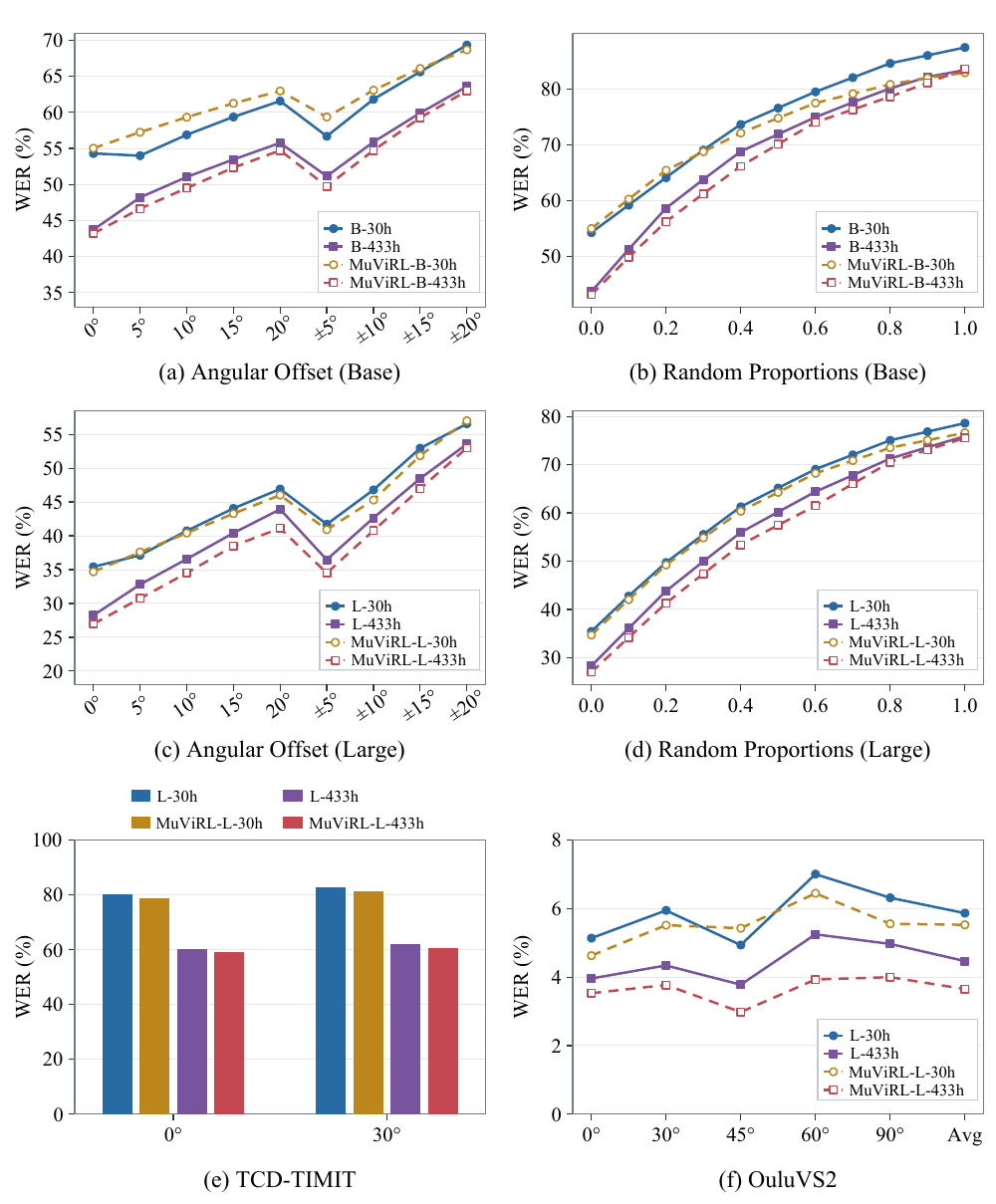}
\caption{Quantitative evaluation on multiview data. B-30h/B-433h denote AV-HuBERT Base pretrained on 433\,h and fine-tuned on 30\,h/433\,h, respectively. L-30h/L-433h denote AV-HuBERT Large pretrained on 1,759\,h and fine-tuned on 30\,h/433\,h, respectively. MuViRL uses the corresponding view augmented data. (a,c) Angular Offset: performance as the range of synthesized yaw offsets expands. (b,d) Random Proportions: performance with varying proportions of synthesized test clips. (e,f) Results on TCD-TIMIT and OuluVS2, respectively.}\label{fig:curves}
\end{figure}

\textbf{Quantitative evaluation.} Fig.~\ref{fig:curves} presents results on real and synthesized multiview data. After OuluVS2 fine-tuning from matched VA-1759h pretraining and VA-433h recognition checkpoints, MuViRL-L achieves 3.65\% overall WER on OuluVS2, compared with 4.47\% for the baseline, an 18.3\% relative improvement. WER is also lower at $60^\circ$ and $90^\circ$, and the overall improvement remains with VA-30h fine-tuning. Direct transfer to TCD-TIMIT uses all 5,782 paired utterances from 59 volunteers with identical preprocessing and decoding. Overall WER is 79.85\% and 59.78\% with 30\,h and 433\,h fine-tuning, compared with 81.48\% and 61.06\% for the baseline. Lower WER at both viewpoints without additional training on the transfer corpus indicates generalization to real multiview data. Evaluation on synthesized data covers different angular offsets and proportions of synthesized test clips. With 433\,h fine-tuning, both Base and Large configurations outperform the baseline in most settings. The recognition gains extend across different angular ranges and proportions of synthesized clips.

\begin{figure}[!t]
\centering\includegraphics[width=1\textwidth]{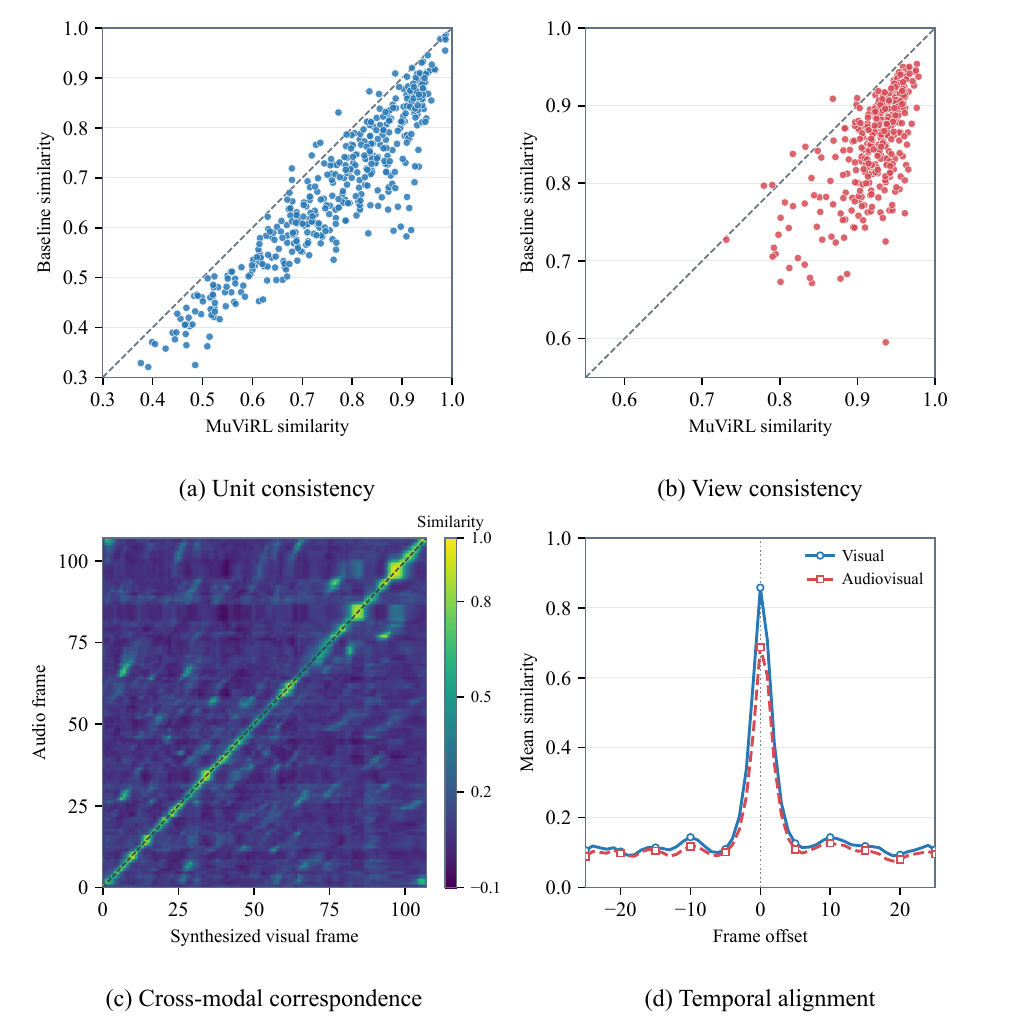}
\caption{Multimodal representation structure. (a) Unit consistency and (b) view consistency on OuluVS2. Points below the diagonal indicate higher similarity with MuViRL. (c) Correspondence between audio and synthesized visual frames on LRS3. The bright diagonal indicates matching timestamps. (d) Mean similarity at different frame offsets for visual and audiovisual pairs from the same sequence. Peaks at zero offset indicate temporal alignment.}\label{fig:features}
\end{figure}

\textbf{Representation structure analysis.} We assess whether representations remain consistent within multimodal units and across synchronized views while preserving temporal distinctions. Table~\ref{tab:representations} presents prediction and retrieval results, and Fig.~\ref{fig:features} shows feature structure. Experiments on both datasets use the Base configurations. Baseline continues prediction on the original view, Random view samples one view per update, and MuViRL uses MMP, Adaptive Weighting, Agg2S, and Temporal RSA without decision or depth weighting.

Prediction accuracy is the proportion of valid masked positions with a correctly predicted unit. Both retrieval metrics~\cite{lu2023retrieval,araujo2025sync} report Recall@1. For each candidate view, cosine similarity ranks its valid frames within the same utterance. Cross-view Recall uses original visual queries, and Cross-modal Recall uses audio queries. A query is correct when the highest-ranked frame has the same timestamp. As shown in Table~\ref{tab:representations}, Cross-view Recall improves from 64.59\% to 67.64\% on LRS3 and from 83.40\% to 88.66\% on OuluVS2. OuluVS2 Cross-modal Recall increases from 64.02\% to 68.60\%. The gains in both recalls indicate more accurate temporal matching across viewpoints and modalities.

Unit consistency averages cosine similarity between distinct normalized features sharing a teacher label. View consistency averages it over synchronized feature pairs within each utterance. MuViRL improves similarity for 436 of 447 units and 353 of 360 test utterances in Fig.~\ref{fig:features}(a,b). Across three seeds, the respective means increase from 0.6379 to 0.6997 and from 0.8514 to 0.9202. Unit centers also become more separated, with squared distance increasing from 1.0403 to 1.1300. For the same LRS3 sequence, the bright diagonal and peaks at zero offset in Fig.~\ref{fig:features}(c,d) indicate temporal alignment. Offsets are synthesized visual frame indices minus original visual or audio frame indices. Supplementary Table~S2 reports prediction and retrieval results for component comparisons.

\subsection{Evaluation of Calibrated View Aggregation}\label{sec:decisions}
\begin{figure}[!t]
\centering\includegraphics[width=1\textwidth]{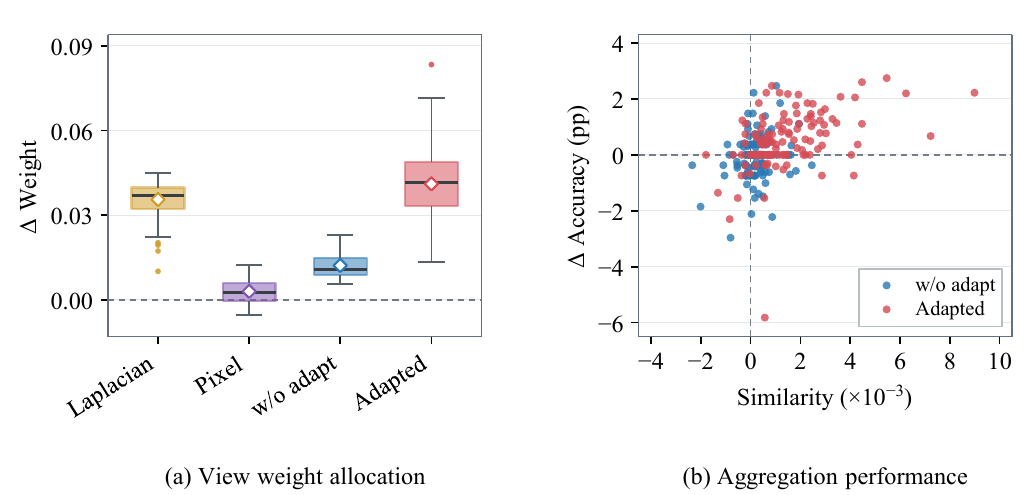}
\caption{View aggregation analysis. (a) Weight gains for views without added distortion. Boxes show quartiles and medians, and diamonds mark means. (b) Changes in similarity and accuracy per utterance relative to aggregation without reliability weighting. Similarity on the horizontal axis is the paired change, scaled by $10^3$. Points in the upper right indicate improved stability and prediction.}\label{fig:decisions}
\end{figure}

To assess how Calibrated View Aggregation supports multimodal representation learning, we evaluate aggregate prediction and recognition from individual views. Table~\ref{tab:decisions} presents aggregation results with the encoder and Adaptive Weighting fixed, and Fig.~\ref{fig:decisions} shows weight allocation and prediction.

\textbf{Decision model adaptation analysis.} The aggregation comparison uses scores from a decision model fine-tuned with four quality grades on ORL, Yale, Notting Hill, and MEAD. Training uses blur and occlusion, while evaluation on pixelation tests generalization. Actors and identities are disjoint across adaptation, calibration, and evaluation. Supplementary Table~S1 presents grading accuracy across the four datasets. Aggregation evaluation uses 120 OuluVS2 test utterances covering all twelve test speakers. Distortion is applied to two oblique views at one of three levels.

\begin{table}[!t]
\caption{Aggregation evaluation with different input configurations. Original inputs are evaluated without added distortion. Adaptation means fine-tuning with four quality grades. w/o reliability means $q_{i,t}=1$. Similarity means cosine agreement with the reference aggregate computed from original inputs, with larger values indicating greater stability.}\label{tab:decisions}
\centering\small\setlength{\tabcolsep}{2pt}
\begin{tabularx}{\textwidth}{@{}Xccccc@{}}
\toprule
\multirow{2}{*}[-1em]{Configuration} & \multirow{2}{*}[-1em]{Adaptation} & \multicolumn{3}{c}{Prediction accuracy (\%) $\uparrow$} & \multirow{2}{*}[-1em]{Similarity $\uparrow$}\\
\cmidrule(lr){3-5}
& & \begin{tabular}[c]{@{}c@{}}Original\\inputs\end{tabular} & \begin{tabular}[c]{@{}c@{}}Blur/\\occlusion\end{tabular} & \begin{tabular}[c]{@{}c@{}}Pixelation\end{tabular} & \\
\midrule
\raisebox{-0.35\height}{\shortstack[l]{w/o reliability\\weighting}} & -- & 37.29 & 36.23 & 37.05 & 0.9702 \\
Laplacian & -- & 37.32 & 36.73 & 37.05 & 0.9723 \\
Pixel similarity & -- & 37.35 & 36.73 & \textbf{37.11} & 0.9854 \\
\midrule
Decision model & $\times$ & 37.44 & 36.82 & 36.87 & 0.9918 \\
Decision model & $\checkmark$ & \textbf{38.22} & \textbf{37.87} & 36.93 & \textbf{0.9926} \\
\bottomrule
\end{tabularx}
\end{table}

\textbf{View aggregation analysis.} As shown in Table~\ref{tab:decisions}, aggregation with the adapted decision model achieves 37.87\% accuracy under blur and occlusion, compared with 36.82\% without adaptation and 36.23\% without reliability weighting. It also gives the highest accuracy on original inputs (38.22\%) and similarity to the reference aggregate (0.9926). The reference uses inputs without added distortion and uniform reliability. Table~\ref{tab:decisions} pools frame contributions across distortion conditions within each seed, then averages over three seeds. Fig.~\ref{fig:decisions} shows weight allocation and paired changes per utterance, averaged over three distortion types, three severity levels, and three seeds. Under unseen pixelation, pixel similarity gives the highest accuracy (37.11\%), compared with 36.93\% using adapted decisions. The adapted decision scores improve aggregate prediction under blur and occlusion.

\textbf{Evaluation of decision scores.} Supplementary Table~S3 evaluates teacher supervision, ranking, and temperature calibration under matched data, initialization, and training budgets. Ranking improves pairwise accuracy from 52.90\% to 54.02\%, while calibration lowers expected calibration error from 0.68\% to 0.49\%. We compare aggregation using uniform reliability, ranking scores, and calibrated scores on the same OuluVS2 split. Base encoder initialization, view sampling, training budgets, Agg2S, and uniformly weighted Temporal RSA remain fixed. Across three seeds, mean WER is 17.52\% with uniform reliability, 17.10\% with ranking scores, and 16.48\% with calibrated scores. Calibrated scores also yield the highest aggregate accuracy (37.56\%). These results show that calibrated decision scores improve aggregate prediction and reduce recognition errors.

\subsection{Evaluation of Aggregation-to-Single Distillation}\label{sec:branches}
To evaluate Agg2S, we compare prediction objectives and aggregation methods in Table~\ref{tab:branches}. Both comparisons reuse the LRS3 split and Base training settings from Section~\ref{sec:representations}, with uniform reliability and Temporal RSA disabled. Adaptive Weighting is used for the aggregation branch in the objective comparison. The aggregation comparison retains MMP and Agg2S.

\begin{table}[t]
\centering
\begin{minipage}{0.8\linewidth}
\caption{Evaluation of Aggregation-to-Single Distillation (Agg2S). $\mathcal L_{\mathrm{single}}$ and $\mathcal L_{\mathrm{agg}}$ mean target prediction from one view and aggregated views, respectively.}\label{tab:branches}
\centering\small\setlength{\tabcolsep}{7pt}
\begin{tabular*}{\linewidth}{@{\extracolsep{\fill}}lcc@{}}
\toprule
Configuration&Cross-view Recall (\%) $\uparrow$&WER (\%) $\downarrow$\\
\midrule
\rowcolor[gray]{0.9}\multicolumn{3}{c}{\textit{Prediction objective}}\\
$\mathcal L_{\mathrm{single}}$&65.32&69.50\\
$\mathcal L_{\mathrm{agg}}$&64.42&68.84\\
MMP&65.25&69.50\\
MMP + Agg2S&\textbf{67.63}&\textbf{66.21}\\
\midrule
\rowcolor[gray]{0.9}\multicolumn{3}{c}{\textit{Aggregation method}}\\
Mean&66.18&69.47\\
Learned&67.21&69.55\\
Cross-view Attention&67.32&69.43\\
Adaptive Weighting&\textbf{67.63}&\textbf{66.21}\\
\bottomrule
\end{tabular*}
\end{minipage}
\end{table}

\textbf{Prediction objective comparison.} MMP with Agg2S achieves 67.63\% Cross-view Recall and 66.21\% WER, compared with 65.25\% and 69.50\% for MMP alone. The OuluVS2 comparison reuses the representation experiment's split and training settings, with uniform reliability and Temporal RSA disabled. Increasing the Agg2S coefficient from zero to 0.1 raises mean Cross-view Recall from 86.73\% to 86.91\% over three seeds (Supplementary Table~S2). The retrieval gains on both datasets and lower LRS3 WER support transferring aggregated predictions to individual views.

\textbf{Aggregation method comparison.} Mean assigns equal weights to all views. Learned estimates weights from individual views, while Cross-view Attention models interactions across views. Adaptive Weighting achieves the highest recall and lowest WER. Cross-view Recall increases from 66.18\% with Mean to 67.63\% with Adaptive Weighting. With the distillation objective fixed, Adaptive Weighting improves temporal correspondence and reduces recognition errors.

\subsection{Ablation Study}\label{sec:ablation}
We conduct an ablation study of MuViRL using the LRS3 dataset. As shown in Table~\ref{tab:ablation}, all configurations use the same data, initialization, and training settings. Baseline uses MMP and Adaptive Weighting with uniform aggregation reliability. Decision and Depth apply utility $u_{i,b}$ and geometry $g_{i,b}$ to RSA weights.

\begin{table}[t]
	\caption{Ablation studies of MuViRL on LRS3. Decision and Depth mean decision and geometric weighting of Temporal RSA, respectively.}\label{tab:ablation}
	\centering\small\setlength{\tabcolsep}{3pt}
	\begin{tabularx}{\textwidth}{@{}cccc>{\centering\arraybackslash}X>{\centering\arraybackslash}X>{\centering\arraybackslash}X@{}}
		\toprule
		\textbf{Agg2S}&\textbf{RSA}&\textbf{Decision}&\textbf{Depth}&\begin{tabular}[c]{@{}c@{}}\textbf{Cross-view}\\\textbf{Recall (\%) $\uparrow$}\end{tabular}&\begin{tabular}[c]{@{}c@{}}\textbf{Cross-modal}\\\textbf{Recall (\%) $\uparrow$}\end{tabular}&\textbf{WER (\%) $\downarrow$}\\
		\midrule
		\multicolumn{4}{c}{Baseline}&65.25&53.93&69.50\\
		\midrule
		$\checkmark$&--&--&--&67.63&55.12&66.21\\
		--&$\checkmark$&--&--&66.22&54.30&68.54\\
		$\checkmark$&$\checkmark$&--&--&67.64&55.32&65.87\\
		$\checkmark$&$\checkmark$&$\checkmark$&--&68.17&56.03&65.07\\
		$\checkmark$&$\checkmark$&--&$\checkmark$&67.84&55.35&65.95\\
		$\checkmark$&$\checkmark$&$\checkmark$&$\checkmark$&\textbf{68.69}&\textbf{56.40}&\textbf{64.75}\\
		\bottomrule
	\end{tabularx}
\end{table}

\textbf{Agg2S.} As shown in Table~\ref{tab:ablation}, Agg2S improves Cross-view Recall from 65.25\% to 67.63\% and Cross-modal Recall from 53.93\% to 55.12\%. WER falls from 69.50\% to 66.21\%. With Temporal RSA retained, Agg2S further improves WER from 68.54\% to 65.87\%. This improvement indicates that prediction distillation provides supervision beyond temporal feature alignment.

\textbf{Reliability-Aware Real--Synthetic Alignment.} Temporal RSA alone achieves 66.22\% Cross-view Recall, 54.30\% Cross-modal Recall, and 68.54\% WER, improving all three metrics over Baseline. When combined with Agg2S, WER improves from 66.21\% to 65.87\%, with recalls of 67.64\% and 55.32\%. The joint window and relation objective improves recognition when used with distillation. With decision weighting of Temporal RSA, WER falls from 65.87\% to 65.07\%, while both recalls increase to 68.17\% and 56.03\%. Depth weighting alone achieves 67.84\% and 55.35\% recall but gives 65.95\% WER. Combining decision and depth weighting yields 68.69\% and 56.40\% recall and 64.75\% WER. Weighting alignment by both predictive usefulness and geometric consistency improves correspondence across views and modalities.

Compared with Baseline, Agg2S with weighted Temporal RSA increases Cross-view Recall from 65.25\% to 68.69\% and Cross-modal Recall from 53.93\% to 56.40\%. WER decreases from 69.50\% to 64.75\%. With masked prediction supervision held fixed, these gains support the combined use of Agg2S and reliability weighted temporal alignment for learning consistent and discriminative representations.

\FloatBarrier
\section{Conclusion}\label{sec:conclusion}
In this paper, we proposed MuViRL, a self-supervised framework for multimodal view-invariant visual speech representation learning. Synchronized 3DGS synthesis provides corresponding views for learning from shared audiovisual targets. Calibrated View Aggregation uses calibrated multimodal decisions and Adaptive Weighting to select informative visual features. Agg2S transfers aggregated predictions to individual views, and Temporal RSA preserves temporal correspondence between original and synthesized representations. Experiments show improved multimodal representation consistency and visual speech recognition under view variation, including generalization to real multiview data. Ablation studies support the contributions of Agg2S and reliability weighted alignment.

\FloatBarrier
\backmatter

\section*{Declarations of Conflict of Interest}
The authors declare that they have no financial or nonfinancial conflicts of interest related to this work.


\bibliographystyle{mir-numeric}
\bibliography{references}

\end{document}